\documentclass[letterpaper, 10 pt, conference]{ieeeconf}  

\IEEEoverridecommandlockouts                              

\usepackage{cite}
\usepackage{amsmath,amssymb,amsfonts}
\usepackage{multirow}
\usepackage{float}
\usepackage{subfigure}
\usepackage{comment}
\usepackage{algorithm}
\usepackage{algpseudocode}
\usepackage{booktabs} 
\usepackage{array}    
\usepackage{siunitx}  
\usepackage{graphicx} 
\usepackage[table]{xcolor} 
\usepackage{hyperref}

\hypersetup{
    colorlinks=true,    
    linkcolor=blue,     
    filecolor=magenta,  
    urlcolor=cyan,      
    citecolor=green,    
}

\title{\textit{C}$^{2}$ROPE: Causal Continuous Rotary Positional Encoding for 3D Large Multimodal-Models Reasoning}

\author{Guanting Ye$^{1}$, Qiyan Zhao$^{2}$, Wenhao Yu$^{3}$, Xiaofeng Zhang$^{2}$, Jianmin Ji$^{4}$, Yanyong Zhang$^{5}$, and Ka-Veng Yuen$^{1,*}$
\thanks{This work is funded by the Science and Technology Development Fund, Macau SAR under Research Grant 001/2024/SKL, the Research Committee of University of Macau under Research Grant CPG2026-00033-FST and the Guangdong-Hong Kong-Macau Joint Laboratory Program under Grant 2020B1212030009.}
\thanks{$^{1}$State Key Laboratory of Internet of Things for Smart City, University of Macau (UM), Macau, China} 
\thanks{$^{2}$Department of Automation, Shanghai Jiaotong University (SJTU), Shanghai, China}
\thanks{$^{3}$Institute of Advanced Technology, University of Science and Technology of China (USTC), Hefei, China}
\thanks{$^{4}$School of Computer Science and Technology, USTC, Hefei, China}
\thanks{$^{5}$School of Artificial Intelligence and Data Science, USTC, Hefei, China}%
\thanks{${*}$ Corresponding author. {\tt\small kvyuen@um.edu.mo}}
}

\begin{document}

\maketitle
\thispagestyle{empty}
\pagestyle{empty}

\begin{abstract}

Recent advances in 3D Large Multimodal Models (LMMs) built on Large Language Models (LLMs) have established the alignment of 3D visual features with LLM representations as the dominant paradigm. However, the inherited Rotary Position Embedding (RoPE) introduces limitations for multimodal processing. Specifically, applying 1D temporal positional indices disrupts the continuity of visual features along the column dimension, resulting in spatial locality loss. Moreover, RoPE follows the prior that temporally closer image tokens are more causally related, leading to long-term decay in attention allocation and causing the model to progressively neglect earlier visual tokens as the sequence length increases. To address these issues, we propose \textbf{\textit{C}$^{2}$RoPE}, an improved \textbf{RoPE} that explicitly models local spatial \textbf{\textit{C}}ontinuity and spatial \textbf{\textit{C}}ausal relationships for visual processing. \textit{C}$^{2}$RoPE introduces a spatio-temporal continuous positional embedding mechanism for visual tokens. It first integrates 1D temporal positions with Cartesian-based spatial coordinates to construct a triplet hybrid positional index, and then employs a frequency allocation strategy to encode spatio-temporal positional information across the three index components. Additionally, we introduce Chebyshev Causal Masking, which determines causal dependencies by computing the Chebyshev distance of image tokens in 2D space. Evaluation results across various benchmarks, including 3D scene reasoning and 3D visual question answering, demonstrate C$^{2}$RoPE's effectiveness. The code is be available at https://github.com/ErikZ719/C2RoPE.

\end{abstract}

\begin{figure}[t]
\centerline{\includegraphics[height=8.5cm]{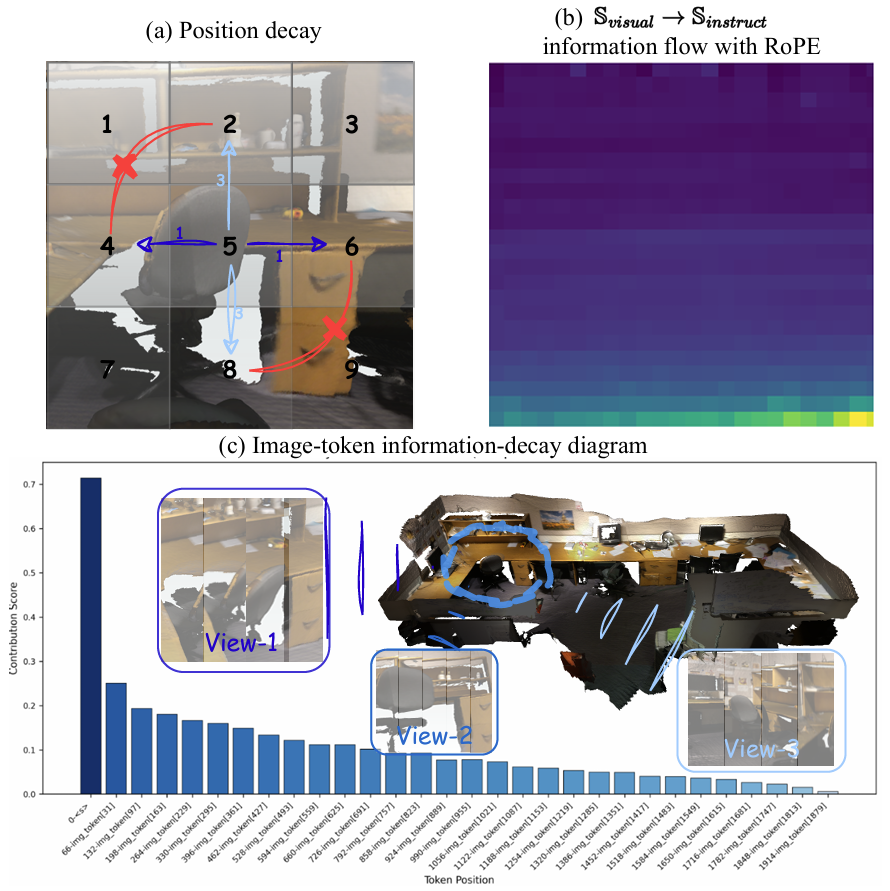}}
\caption{(a) illustrates RoPE’s raster-scan positional index assignment; (b) visualizes the information flow from image tokens to instruction tokens, where flows from each image token are aggregated and arranged according to their positions in 2D space. The statistics are averaged over a 3K VQA subset. (c) shows the quantitative analysis of information flow from visual tokens to output tokens.}
\label{Analysis}
\vspace{-0.5cm}

\end{figure}

\setcounter{secnumdepth}{2}

\section{INTRODUCTION}

Building 3D Large Multimodal Models (LMMs)~\cite{zhu2024llava, deng20253d} has emerged as a promising research direction with significant potential for applications such as autonomous robotics and navigation systems. With the rapid advancement of Large Language Models (LLMs)~\cite{yang2025qwen3}, researchers are increasingly focusing on leveraging their impressive generation capabilities to enhance the understanding of 3D scene~\cite{ll3da, ross3d}.

Most mainstream 3D LMMs adopt a visual information injection paradigm similar to that of 2D LMMs~\cite{liu2024improved, qwen2-vl, zhang2024omg, team2023gemini}, aligning visual representations with LLM embedding spaces to enable cross-modal reasoning and instruction-conditioned policy synthesis for robotic navigation, manipulation, and human-robot interaction~\cite{yu2024pathrl, duan2025stdarm, yu2024ldp}. To construct effective 3D visual representations, some approaches project 3D points into 2D visual representations~\cite{hong20233d, zhi2024lscenellm}, while others directly encode 3D features~\cite{chat-3d, xu2024pointllm}. More recently, LLaVA-3D ~\cite{zhu2024llava} has innovatively embedded point cloud features into multi-view 2D image tokens, resulting in 3D perception-enhanced visual tokens. In these models, visual features are typically encoded as tokens and concatenated with text tokens to create a multi-model input sequence, which is then processed autoregressively by LLMs. While this paradigm allows the 3D LMMs to be more compatible with LLM architectures, the implications of inheriting intrinsic mechanisms originally designed for natural language processing remain underexplored.

Recent research on information flow provides an intuitive interpretability approach for analyzing the intrinsic mechanisms of black-box models ~\cite{Wang2023LabelWA, chen2024image}. Building on this, prior studies ~\cite{cca-llava, mca, Zhu2025MitigatingOH} have shown through image-to-instruction information flow analysis that Rotary Position Encoding (RoPE) ~\cite{su2024roformer}, a core component for positional modeling in LLMs, limits effective interaction between images and instructions in 2D LMMs. This raises the question:

\noindent
\textbf{\textit{For 3D LMMs, does inheriting RoPE from LLMs also introduce limitations?}}

In RoPE, image tokens are assigned unique temporal positional indices following a raster-scan order (left to right, top to bottom). However, this indexing scheme disrupts the continuity of visual features along the vertical (column) dimension. As illustrated in Fig.~\ref{Analysis} (a), the positional indices of spatially adjacent image tokens are not continuous, a phenomenon we refer to as \textbf{spatial locality loss}.

Moreover, since RoPE treats temporally closer tokens as more causally relevant, it induces a long-term decay in attention allocation. To investigate information interaction under long-term decay, we visualize the image-to-instruction information flow . As shown in Fig.\ref{Analysis} (b), although visual tokens in 3D LMMs carry rich 3D information, only a small subset near the end of the sequence receives high attention, while most contribute only sparse information flow. We refer to this phenomenon as \textbf{visual tokens neglect}. Furthermore, Fig.\ref{Analysis} (c) shows the quantitative analysis of information flow from image tokens to output tokens, revealing that this neglect persists during response generation and becomes more pronounced as multi-view images further extend the sequence. A detailed discussion of spatial locality loss and visual tokens neglect is provided in Section 3.



Based on the above analysis, we propose a novel positional encoding method, \textbf{\textit{C}$^{2}$RoPE}, designed to preserve the local spatial \textbf{\textit{C}}ontinuity of visual tokens and capture spatial \textbf{\textit{C}}ausal relationships. To achieve this, we introduce a spatio-temporal continuous positional modeling mechanism. First, the 2D image is projected onto a Cartesian coordinate system to obtain each token’s spatial coordinates $(x,y)$. These coordinates are then combined with RoPE’s original 1D temporal index $m$ to form a triplet hybrid positional index $(m,x,y)$. Finally, we encode spatio-temporal information by assigning distinct frequency bands to each component in $(m,x,y)$. Additionally, we introduce Chebyshev Causal Masking, which determines the causal relationships of visual tokens based on their relative positional distance from the origin in the Cartesian coordinate system. This masking is applied to the self-attention matrix during decoding to alleviate image token neglect caused by long-term decay.


To fully evaluate the performance of \textit{C}$^{2}$RoPE, we conduct experiments on benchmarks including 3D scene reasoning and 3D visual question answering, demonstrating consistent performance improvements over the baselines. Compared to LLaVA-3D, our method achieves an improvement of +4.3 on EM@1 on ScanQA, and improvements of +1.2 on EM@1 and +1.2 on EM@R on SQA3D. Our main contributions are summarized as follows:

\begin{itemize}
\item This paper conducts an in-depth analysis of the limitations of applying RoPE in 3D LMMs, primarily manifested as spatial locality loss and visual tokens neglect.

\item Motivated by our analysis, we propose \textit{C}$^{2}$RoPE, which incorporates two key designs: a spatio-temporal continuous positional embedding mechanism and a Chebyshev causal masking strategy.

\item Experiments on multiple benchmarks and across different baselines support the efficacy of our design.

\end{itemize}

\section{Related Work}

\subsection{3D Large Multimodal Models}

Large Language Models (LLMs) have made significant progress in recent years, demonstrating impressive performance across various scenarios ~\cite{achiam2023gpt, chowdhery2023palm, dai2023instructblip, li2023evaluating,wang2025largerlanguagemodelsimply}. These achievements have driven a growing number of researchers to develop Large Multimodal Models (LMMs) with 3D scene understanding capabilities ~\cite{chen2023minigpt,scan2cap, scannet, leo, zhang2024omg}. Current 3D LMMs typically use a pre-trained LLMs ~\cite{vicuna} as the decoder, and then achieve cross-modal feature alignment by projecting image representations into the LLM space and jointly decoding them with text tokens. Built on this basic architecture and the cross-modal alignment paradigm, subsequent research has primarily focused on constructing more effective 3D visual representations. For instance, ScanReason ~\cite{ScanReason}, Pointllm ~\cite{xu2024pointllm} and Ll3da ~\cite{ll3da} leverage visual encoders to directly encode 3D point clouds features. Video-3D LLM ~\cite{video-3d-llm}, Scene-LLM ~\cite{scene-LLM}, and 3D LLM ~\cite{hong20233d} extract 3D visual representations from multi-view image inputs. The recently proposed LLaVA-3D ~\cite{zhu2024llava} embeds 3D point cloud features into multi-view image tokens, resulting in 2D image tokens enhanced with 3D awareness. However, these 3D LMMs inherits the LLM's positional modeling scheme, which can lead to issues such as “spatial locality loss” or the “image tokens neglect” problem.

\subsection{Position Encoding Schemes in LMMs}

Most 3D LMMs use position encoding methods inherited from LLMs, primarily RoPE ~\cite{su2024roformer, gao2024tc, agrawal2024pixtral}. In their approach, images are flattened into a 1D token sequence, with positional indices assigned to each image token in a raster-scan order. Subsequently, RoPE introduces complex rotations into the token embedding space. Compared to absolute ~\cite{Vaswani2017AttentionIA} or relative position encoding ~\cite{ke2021rethinking}, RoPE provides a smoother inductive bias for interactions between nearby and distant tokens. Subsequent works have proposed corresponding improvements to address the limitations of RoPE in specific domains. HoPE ~\cite{li2025hope} proposes a hybrid frequency allocation strategy to improve the long-context capabilities. CoMemo ~\cite{liu2025comemo} and MiniCPM-V ~\cite{Yao2024MiniCPMVAG} proposed different RoPE variants to alleviate remote decay in DHR scenarios. CCA ~\cite{cca-llava} and MCA ~\cite{mca} proposed different heuristic positional index assignment strategies to mitigate hallucinations caused by long-term decay in 2D LMMs. To enhance multimodal processing, VideoRoPE ~\cite{wei2025videorope} and M-RoPE ~\cite{qwen2-vl} decomposing positional embedding into 1D, 2D, and 3D components and set different frequency distributions, effectively improving the performance of LMMs in video scenes. Although the aforementioned RoPE improvements have enhanced performance in various scenarios, research has largely focused on 2D LMMs. Positional encoding in 3D LMMs remains underexplored. To the best of our knowledge, this work is the first to provide an in-depth analysis of RoPE's limitations in the context of 3D LMMs.


\section{Analysis and Motivation}

In this section, we further examine the spatial locality loss and visual tokens neglect, and conduct pilot experiments to investigate their impact. We begin by introducing the architecture of LLaVA-3D ~\cite{zhu2024llava}, the 3D LMMs we use as our baseline, and the mechanism of RoPE. Next, we elaborate on the spatial locality loss induced by RoPE’s positional indices, which fail to encode information along the column dimension. Finally, we present visualizations of image-to-instruction information flow, along with a quantitative analysis of information flow from image tokens to output token. These results reveal the image tokens neglect phenomenon. The above findings serve as the motivation and foundation for our proposed design.

\subsection{Preliminary}

\noindent
\textbf{3D LMMs:} We adopt LLaVA-3D \cite{zhu2024llava} as baseline, which consists of a pretrained image encoder $F_v$, a large language model $F_t$, and a projector $f$. Given multi-view 2D images $I_v \in \mathbb{R}^{V\times 3 \times H \times W}$, where $V$ denotes the number of views. The image encoder $F_v$ processes $I_v$ into a multi-view 2D patch feature sequence $\mathbb{S}_{2d} \in \mathbb{R}^{V\times d \times h \times w}$, where $d$ denotes the embedding dimension. LLaVA-3D then injects 3D positional embeddings $P_{3d} \in \mathbb{R}^{V\times 3 \times h \times w}$, encoded from point cloud features using a 2-layer MLP, into $\mathbb{S}_{2d}$, resulting in a 3D geometry-aware patch feature sequence $\mathbb{S}_{3d}  = \mathbb{S}_{2d} + P_{3d} $. $\mathbb{S}_{3d} \in \mathbb{R}^{V\times d \times h \times w}$ are then sequentially sent into the projector $f$ to obtain the final visual token sequence $\mathbb{S}_{v} = f(\mathbb{S}_{3d})= \{ w_m \}^v_{m=1}$. The instruction text $I_t$ is encoded by $F_t$ into the text tokens sequence $\mathbb{S}_{t} = F_t(I_t)= \{ w_m \}^t_{m=1}$. Here, $v$ and $t$ denote the lengths of the visual and textual token sequences, respectively. Both $\mathbb{S}_{v}$ and $\mathbb{S}_{t}$ share the same embedding dimension $d$, and are concatenated to form the final multimodal input tokens sequence $\mathbb{S}= Concat( \mathbb{S}_{v}, \mathbb{S}_{t} ) = \{w_m, m \in [1, v+t] \} $.

\noindent
\textbf{RoPE in LLaVA-3D:} LLaVA-3D inherits RoPE from LLMs to encode the positional information of query and key tokens. Given a query token $Q$ and a key token $K$ with $d$ dimensions, which are derived from $\mathbb{S}$. RoPE divides the dimension $d$ into $d/2$ pairs, with each pair assigned predefined sinusoidal function values given by $\left\{ \theta_i = 10000^{-2(i-1)/d} \right\}$, $\quad i \in ( 1, 2, \dots, d/2)$. Subsequently, the rotation matrix $r^{(i)}$ is constructed as: 
\begin{equation}r^{(i)} = 
\begin{pmatrix}
  \cos(\theta_i) & -\sin(\theta_i) \\
  \sin(\theta_i) & \cos(\theta_i) \\
\end{pmatrix}
\end{equation} Each pairwise rotation matrix $r^{(i)}$ is concatenated along the diagonal to form the overall rotation matrix, denoted as $R_{m} = \mathrm{diag}(r^{(1)}, \dots, r^{(d/2)})$, where $m$ denotes the position index of each input token $ w_m $. RoPE then applies rotation matrix $R_{m}$ to self-attention $A_{n, m}$ between the n-th query $Q_n$ and m-th key $K_m$ as: 
\begin{equation}
\resizebox{\columnwidth}{!}{
$ A_{n, m} = \text{softmax}\left( \frac{(Q_nR_n)(K_mR_m)^T}{\sqrt{d}} \right) = \text{softmax}\left( \frac{Q_nR_{n-m}K^T_m}{\sqrt{d}} \right)
$ }
\end{equation}
The rotation matrix $R_{n-m}$ is formally given as follows:
\begin{equation}
\resizebox{\columnwidth}{!}{
  $\displaystyle R_{n-m} = \begin{pmatrix}
  \cos\theta_1(n-m) & -\sin\theta_1(n-m)  & \dots & 0 & 0\\
  \sin\theta_1(n-m) & \cos\theta_1(n-m)  & \dots & 0 & 0\\
  \vdots & \vdots  & \ddots & \vdots & \vdots\\
  0 & 0  & \dots & \cos\theta_{d/2}(n-m) & -\sin\theta_{d/2}(n-m) \\
  0 & 0  & \dots & \sin\theta_{d/2}(n-m) & \cos\theta_{d/2}(n-m)
  \end{pmatrix}$ 
  }
\end{equation}

\subsection{Spatial Locality Loss}

In vanilla RoPE, the input 2D image $I_v$ is flattened into the 1D tokens sequence $\mathbb{S}_{v}= \{ w_m \}^v_{m=1}$, where $m$ denotes the positional index of image tokens, which is assigned in a raster-scan order as follows:
\begin{equation}
\resizebox{\columnwidth}{!}{
  $\left[ 
\begin{array}{ccccccccc}
1 & 2 &  & &\cdots & & &\sqrt{v}-1 & \sqrt{v} \\
\sqrt{v}+1 & \sqrt{v}+2  & & & \cdots & && 2\sqrt{v}-1 &2\sqrt{v}\\
& & \ddots&  & & && & \\
\vdots & \vdots & & \frac{v}{2}-\frac{\sqrt{v}}{2}& &\frac{v}{2}-\frac{\sqrt{v}}{2}+1 &   & \vdots & \vdots \\
\vdots & \vdots &  & \frac{v}{2}+\frac{\sqrt{v}}{2} && \frac{v}{2}+\frac{\sqrt{v}}{2}+1&   & \vdots & \vdots \\
&  & & & & & \ddots& & \\
v-2\sqrt{v}+1 & v-2\sqrt{v}+2 &  & & \cdots& && v-\sqrt{v}-1 & v-\sqrt{v}\\
v-\sqrt{v}+1 & v-\sqrt{v}+2 &  & &\cdots &&& v-1 & v 
\end{array}
\right]$ 
  }
\end{equation}Such image positional indices consider only the temporal positions of image tokens in the 1D input sequence while neglecting their positions in the 2D spatial domain. Specifically, although the sequential ordering of RoPE positional indices implicitly preserves the continuity of image tokens along the row dimension, the row-by-row alignment results in a spatial locality loss of visual features along the column dimension. In Fig.~\ref{Analysis} (a), we illustrate this spatial locality loss more intuitively: along the column dimension, the positional indices of locally adjacent image tokens are not continuous.



\begin{figure*}[t]
\centering
\includegraphics[width=0.9\textwidth]{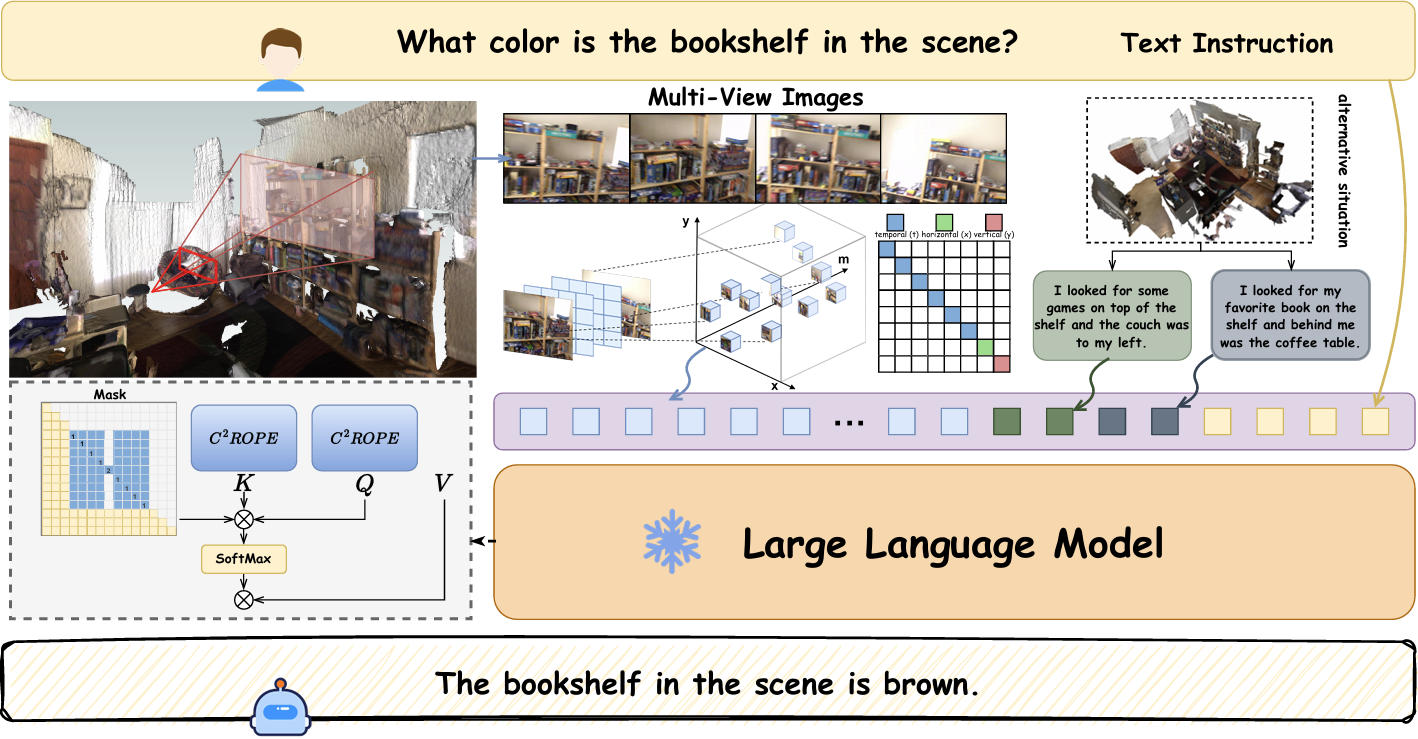}
\caption{Structure of $C^2$RoPE. We integrate temporal indices and Cartesian spatial coordinates to form a triplet hybrid positional index that preserves local continuity of visual tokens. A frequency-based allocation encodes complementary spatio-temporal cues across the triplet, and Chebyshev Causal Masking enforces locality-aware causality by modulating attention with the 2D Chebyshev distance. This design mitigates long-range attention decay and improves 3D scene reasoning in LMMs.}
\label{method}
\vspace{-0.1cm}

\end{figure*}

\subsection{Image Tokens Neglect}

In RoPE, the rotation matrix $R_{n-m}$ incorporates the relative positional distance information $(n - m)$ between each query-key pair. This paradigm inherits the long-term decay property of sinusoidal encoding, where the decrease of $A_{n, m}$ as the $n-m$ increases. This prior is originally designed for natural language modeling, assuming temporally closer tokens are more causally related. In Fig.~\ref{Analysis} (b), we present the visualization of image-to-instruction information flow to examine this property. We observe that attention is significantly concentrated on a small subset of visual tokens near the instruction tokens, which are spatially located in the lower-right region, while the majority are neglected.

We further analyze the quantitative results of information flow from image tokens to output tokens in Fig.~\ref{Analysis} (c) and observe two important phenomena: (1) The initial tokens receive high attention. This is because, in causal self-attention, these tokens act as attention sinks to absorb redundant attention ~\cite{Wang2023LabelWA, Zhang2024SeeingCB}. (2) A large number of image tokens are neglected. The long-term decay of RoPE continues to affect autoregressive next-token prediction, causing visual tokens that are temporally distant from the output tokens to be considered weakly causal and thus assigned minimal attention. Moreover, LLaVA-3D introduces multi-view image inputs to represent 3D spatial information, introducing a critical trade-off: the growing sequence of image tokens exacerbates the image token neglect phenomenon.


\section{Method}

To address these challenges, we propose \textit{C}$^{2}$RoPE, designed to enhance image scene information perception in LLaVA-3D. As shown in Fig.~\ref{method}, \textit{C}$^{2}$RoPE consists of two key designs: (1) a spatio-temporal continuous positional embedding mechanism that explicitly models both the 1D temporal positions and 2D spatial positions of visual tokens to mitigate spatial locality loss; and (2) Chebyshev Causal Masking, which defines causal relationships based on the Chebyshev distance of image tokens in the Cartesian coordinate system, effectively alleviating RoPE’s long-term decay and image token neglect. Details are presented in the following sections.


\begin{figure}[t]
\centerline{\includegraphics[scale=0.33]{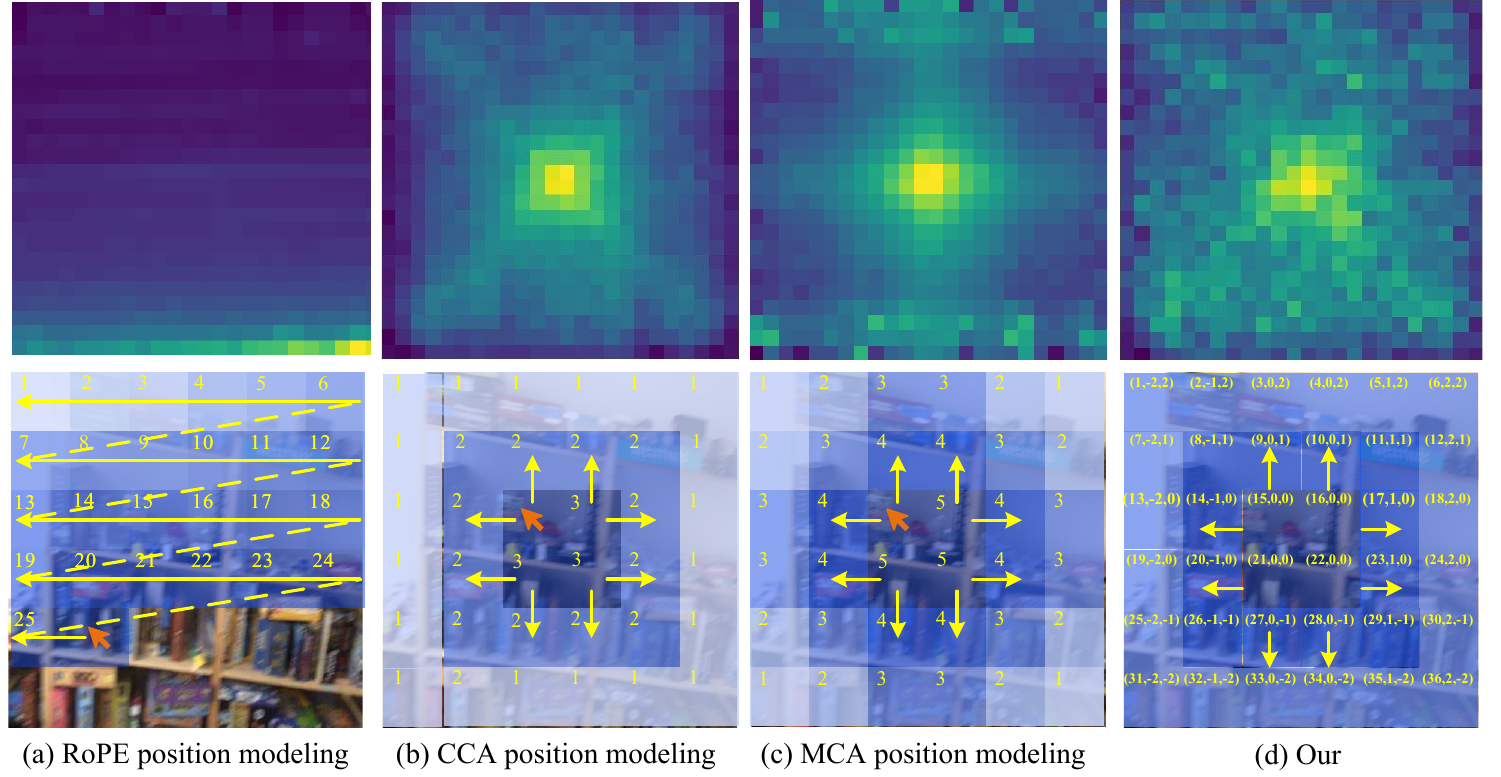}}
\caption{Long-distance position information attenuation of llava-3D, (a) represents the long-distance information attenuation in LLaVA, (b) (c) represent CCA-LLaVA \cite{cca-llava} and MCA-LLaVA \cite{mca} to alleviate the decay of long-distance information by optimizing ROPE, (d) represents our method.}
\label{cca-mca-our}
\vspace{-0.4cm}

\end{figure}

\begin{figure*}[t]
\centering
\includegraphics[width=0.8\textwidth]{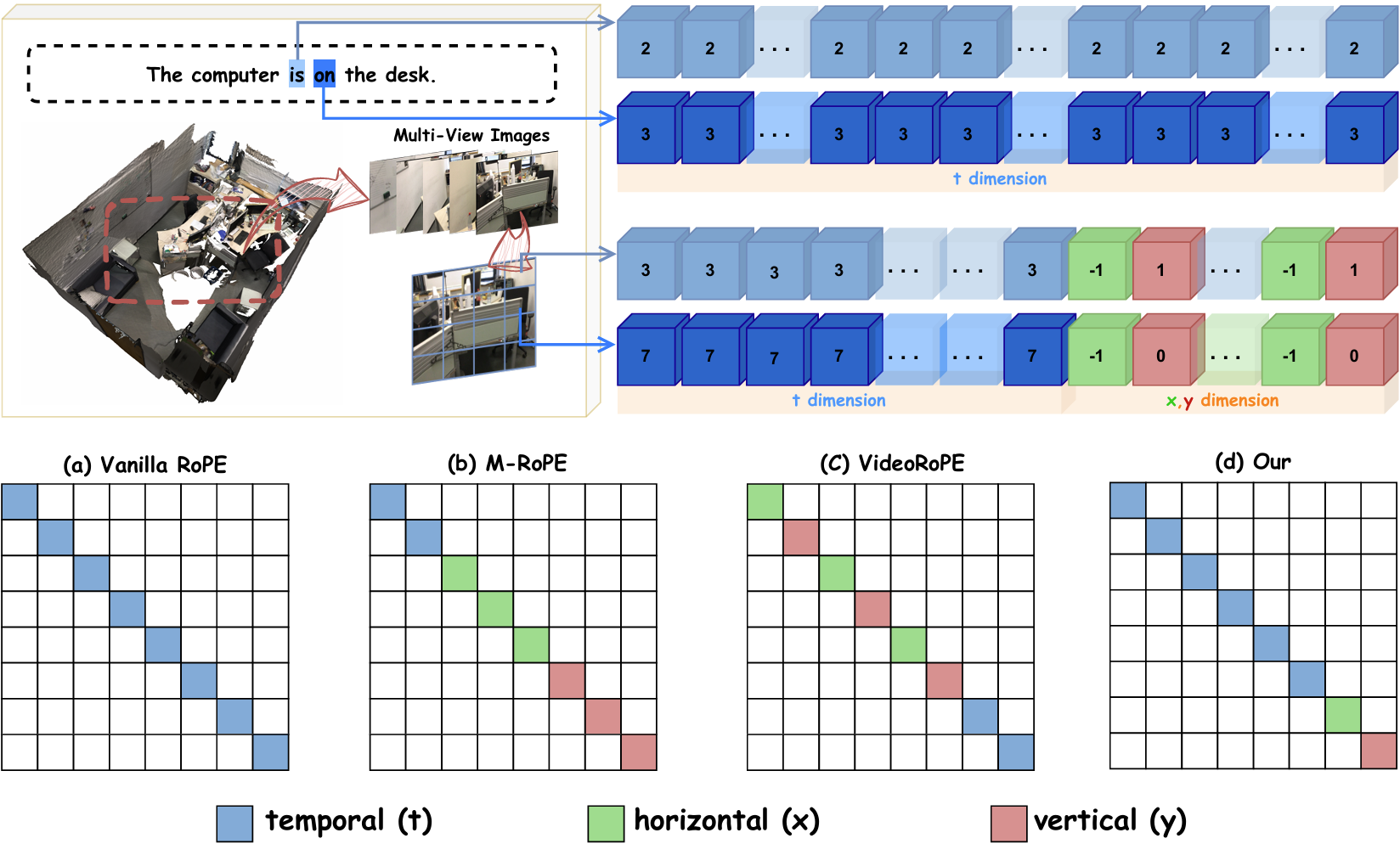}
\caption{The position embeddings of adjacent tokens for Vanilla RoPE (left), frame‑aligned visual tokens for M‑RoPE (center‑left), interleaved spatio‑temporal tokens for VideoRoPE (center‑right), and our $C^2$RoPE (right) with triplet hybrid indexing and Chebyshev causal masking.}
\label{method}
\end{figure*}

\begin{table*}[t]
\centering
\caption{Performance comparison on ScanQA and SQA3D benchmarks. "EM@1" represents top-1 exact match, "B-4" represents "BLEU-4", "MET" represents "METEOR", "RGE" represents "ROUGE", "EM@R" represents refined EM1.}
\footnotesize
\setlength{\tabcolsep}{2.5pt} 
\renewcommand{\arraystretch}{0.85} 
\begin{tabular}{@{}l@{\hspace{6pt}}ccccc@{\hspace{6pt}}cc@{}@{\hspace{6pt}}cc@{}}
\toprule
 & \multicolumn{5}{c}{\textbf{ScanQA}} & \multicolumn{2}{c}{\textbf{SQA3D (test)}} & \multicolumn{2}{c}{\textbf{ScanRefer}} \\
\cmidrule(lr){2-6} \cmidrule(lr){7-8}\cmidrule(lr){9-10}
\textbf{Methods} & 
EM@1$\uparrow$ & 
B-4$\uparrow$ & 
MET$\uparrow$ & 
RGE$\uparrow$ & 
CIDEr$\uparrow$ & 
EM@1$\uparrow$ & 
EM@R$\uparrow$ &
Acc@0.25$\uparrow$ &
Acc@0.5$\uparrow$ \\
\midrule

\textit{Expert Models} & & & & & & & \\
ScanQA \cite{scannet}       & 21.1 & 10.1 & 13.1 & 33.3 & 64.9 & {-} & {-} & {-}& {-}\\
3D-VLP \cite{3d-vlp}        & 21.1 & 11.2 & 13.5 & 34.5 & {-}  & {-} & {-} & {-}& {-}\\
3D-VisTA \cite{3d-vista}    & 22.4 & {-}  & 13.9 & 35.7 & {-}  & {-} & {-} & {-}& {-}\\
\midrule

\textit{2D LLMs} & & & & & & & \\
InternVl2-8B \cite{team2024internvl2}   & {-} & 3.3 & 14.5 & 34.3 & 62.5 & 33.0 & 45.3 & {-}& {-}\\
Qwen2-Vl-7B \cite{qwen2-vl}             & {-} & 3.0 & 11.4 & 29.3 & 53.9 & 40.7 & 46.7 & {-}& {-}\\
LLaVA-Video-7B \cite{llava-next-video}  & {-} & 3.1 & 17.7 & 44.6 & 88.7 & 48.5 & {-} & {-}& {-}\\
\midrule

\textit{3D LLMs} & & & & & & & \\
Chat-3D \cite{chat-3d}         & {-} & 6.4 & 11.9 & 28.5 & 53.2 & {-} & {-} &35.9 &30.4\\
3D-LLM \cite{hong20233d}       & 20.5 & 12.0 & 14.5 & 35.7 & 69.4 & {-} & {-} &30.3 & {-}\\
Scene-LLM \cite{scene-LLM}     & 27.2 & 12.0 & 16.6 & 40.0 & 80.0 & 54.2 & {-} & {-}& {-}\\
LL3DA \cite{ll3da}             & {-} & {-}  & 15.9 & 35.9 & 80.0 & 53.6 & {-} & {-}& {-}\\
LEO \cite{leo}                 & 24.5 & 13.2 & 20.0 & 49.2 & 101.4 & 50.0 & {-} & {-}& {-}\\
ChatScene \cite{zchatscene}    & 21.6 & 14.3 & 18.0 & 41.6 & 87.7 & 54.6 & 57.5 &55.5&50.2\\
Grounded 3D-LLM \cite{grounded3d} & {-} & 13.4 & {-} & 72.7 & {-} & {-} & {-} &47.9&44.1\\
Video-3D-LLM \cite{video-3d-llm} & 30.1 & 16.4 & 20.0 & 49.3 & 102.1 & 58.6 & {-} & {-}& {-}\\
Ross3D \cite{ross3d}           & 30.8 & 17.9 & 20.9 & 50.7 & 107.0 & 63.0 & 65.7 & {-}& {-}\\
AdaToken-3D \cite{iros}        & 30.0 & 13.3 & 17.6 & 44.7 & 91.8 & 54.8 & {-} & {-}& {-}\\
LLaVA-3D \cite{zhu2024llava}   & 27.0 & 14.5 & 20.7 & 50.1 & 91.7 & 55.6 & 53.1 &50.1 &42.7\\
\midrule
Ours                           & \textbf{31.3} & \textbf{23.0} & \textbf{34.1} & \textbf{52.6} & \textbf{109.8} & \textbf{56.8} & \textbf{54.3} & \textbf{50.5}& \textbf{42.9}\\
\bottomrule
\end{tabular}
\label{tab:mme}
\vspace{-2em}
\end{table*}

\subsection{Spatio-temporal Continuous Positional Embedding Mechanism}

\noindent
\textbf{Hybrid Image Positional Indexing:} The analysis in Section 3.2 reveals that RoPE assigns positional indices to image tokens using a raster-scan order, which only represents their temporal positions in the input token sequence but fails to capture their positions in the spatial domain. Although the raster-scan positional indexing implicitly preserves the continuity of image tokens along the row dimension, it disrupts their continuity along the column dimension. As shown in Fig.~\ref{cca-mca-our} (d), we first project the image tokens onto a 2D Cartesian coordinate system to determine the spatial position of each image token. Specifically, the tokens at the image center are treated as the origin, with the axes aligned to rows and columns the positive directions extending rightward and upward, and the coordinates incrementing by 1. We then integrate each token’s $(x,y)$ coordinates with its original RoPE index $m$, forming a triplet hybrid image tokens positional index $(m,x,y)$. The final positional index of image tokens is as follows:
\begin{equation}
\resizebox{\columnwidth}{!}{
  $\left[ 
\begin{array}{ccccccccc}
(1, 1 - \frac{\sqrt{v}}{2},\frac{\sqrt{v}}{2}-1) & (2,2 - \frac{\sqrt{v}}{2},\frac{\sqrt{v}}{2}-1) &  & &\cdots & & &(\sqrt{v}-1, \frac{\sqrt{v}}{2}-2,\frac{\sqrt{v}}{2}-1) & (\sqrt{v}, \frac{\sqrt{v}}{2}-1,\frac{\sqrt{v}}{2}-1) \\
(\sqrt{v}+1, 1 - \frac{\sqrt{v}}{2},\frac{\sqrt{v}}{2}-2) & (\sqrt{v}+2,2 - \frac{\sqrt{v}}{2},\frac{\sqrt{v}}{2}-2)  & & & \cdots & && (2\sqrt{v}-1, \frac{\sqrt{v}}{2}-2,\frac{\sqrt{v}}{2}-2) & (2\sqrt{v}, \frac{\sqrt{v}}{2}-1,\frac{\sqrt{v}}{2}-2)\\
& & \ddots&  & & && & \\
\vdots & \vdots & & (\frac{v-\sqrt{v}}{2}, 0,0) & &(\frac{v-\sqrt{v}}{2}+1, 0,0) &   & \vdots & \vdots \\
\vdots & \vdots &  & (\frac{v+\sqrt{v}}{2}, 0,0) && (\frac{v+\sqrt{v}}{2}+1, 0,0) &   & \vdots & \vdots \\
&  & & & & & \ddots& & \\
(v-2\sqrt{v}+1, 1 - \frac{\sqrt{v}}{2},2 - \frac{\sqrt{v}}{2}) & (v-2\sqrt{v}+2,2 - \frac{\sqrt{v}}{2},2 - \frac{\sqrt{v}}{2}) &  & & \cdots& && (v-\sqrt{v}-1, \frac{\sqrt{v}}{2}-2,2 - \frac{\sqrt{v}}{2}) & (v-\sqrt{v}, \frac{\sqrt{v}}{2}-1,2 - \frac{\sqrt{v}}{2})\\
(v-\sqrt{v}+1, 1 - \frac{\sqrt{v}}{2},1 - \frac{\sqrt{v}}{2}) & (v-\sqrt{v}+2,2 - \frac{\sqrt{v}}{2},1 - \frac{\sqrt{v}}{2}) &  & &\cdots &&& (v-1, \frac{\sqrt{v}}{2}-2,1 - \frac{\sqrt{v}}{2}) & (v,\frac{\sqrt{v}}{2}-1,1 - \frac{\sqrt{v}}{2}) 
\end{array}
\right]$ 
  }
\end{equation}

To compare our design, we illustrate the positional index assignment of RoPE and its variants in Fig.~\ref{cca-mca-our} (a)-(b). Compared to these methods, our approach explicitly models both the temporal and spatial positional information of image tokens. Notably, the temporal component of our positional index reflects the ordering of image tokens within the input sequence, rather than the temporal correlation between adjacent video frames \cite{wei2025videorope, li2025hope}.


\noindent
\textbf{Frequency Allocation Strategy:} After constructing the triplet hybrid image positional index $(m,x,y)$, two further considerations arise:

\noindent
\textit{\textbf{Question1: How to ensure compatibility with the one-dimensional positional indices of text tokens?}}

In our positional index $(m,x,y)$, we retain RoPE’s original index $m$ for image tokens to maintain the absolute positions of image and text tokens in the original input sequence. Additionally, we still use the temporal position $m$ of image and text tokens for relative position calculation, ensuring consistency with the original RoPE when modeling relative positional dependencies. These designs remove the need for any additional processing of text token positional indices.

\noindent
\textit{\textbf{Question2: How to encode temporal and spatial information in hybrid image positional index?}}

Advanced video positional encoding methods, such as M-RoPE~\cite{qwen2-vl} and VideoRoPE~\cite{wei2025videorope}, encode multi-dimensional information by assigning distinct frequency bands to different components of the positional index. As shown in Fig.~\ref{method}, RoPE allocates all 128 frequency dimensions to the temporal index. In contrast, M-RoPE assigns lower-frequency bands to the temporal component and higher-frequency bands to the horizontal and vertical components. VideoRoPE adopts the opposite strategy, prioritizing higher-frequency bands for the temporal component while alternately allocating lower-frequency bands to the horizontal and vertical components.

Notably, the temporal component in previous methods is defined at the image level, whereas our approach operates at the token level. For an image token with positional index $(m, x, y)$ and a rotation matrix of dimension $d = 128$, we interleave $x$ and $y$ in the last 32 dimensions while assigning the remaining 96 dimensions to $m$. This design is driven by two considerations: (1) Lower dimensions correspond to higher frequencies and are highly sensitive to variations. Assigning high frequencies to positional components $x,y$ may cause the model to overfocus on spatial positional changes of image tokens, thereby disrupting the LLMs’ well-trained temporal positional dependencies ~\cite{barbero2025round}; (2) RoPE induces a massive values pattern in query and key tokens, which is essential for capturing the contextual semantics of input tokens ~\cite{jin2025massive}. To preserve this property, we maintain a broad range of frequency bands aligned with RoPE in our allocation strategy.

\subsection{Chebyshev Causal Masking}

Standard RoPE implementations in language modeling assume that temporally closer tokens in text sequences are more causally related. The information flow analysis presented in Section 3.4 reveals that this assumption leads to long-term decay in attention allocation and the neglect of image tokens. Recent studies~\cite{mca, cca-llava} indicate that visual information is inherently structured across a two-dimensional spatial domain, where neighboring image tokens typically exhibit stronger causal relationships. Motivated by this observation, we propose the Chebyshev Causal Masking to explicitly encode this spatial causal relationships. Leveraging the spatial symmetry of images, we set the origin of the Cartesian coordinate system positioned at the image center as the reference point for modeling long-term decay. The causal correlation between tokens is determined by their Chebyshev distance from this origin: the farther a token is, the stronger its attention decay. Tokens sharing the same Chebyshev distance are grouped as correlated within visual self-attention. Fig.~\ref{cca-mca-our} illustrates the key differences between our approach and prior methods: we directly modify the default causal attention masking based on the prior of causal relationships in images. Additionally, we adopt the triplet image positional indexing described in Section 4.1, which preserves the continuity of image tokens in both temporal and spatial dimensions.

\section{Experiments}
\begin{figure}[t]
\centering
\includegraphics[width=0.9\columnwidth]{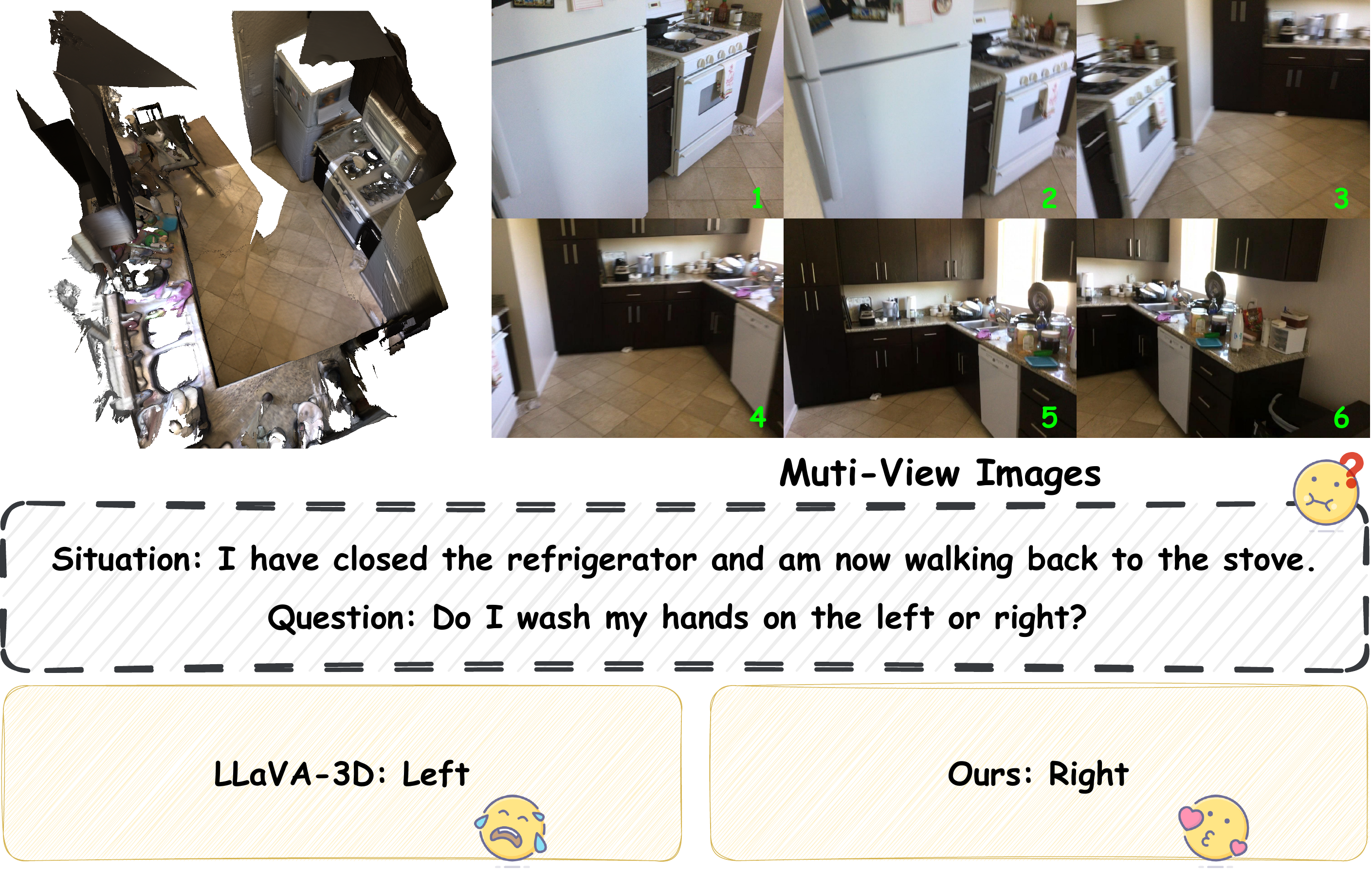}
\caption{The case study is conducted on samples from SQA3D, where we provide identical model inputs to examine the generated responses.}
\label{case}
\end{figure}


\textbf{Experiment models} 
This paper adopts LLaVA-3D-7B ~\cite{zhu2024llava} as the baseline model and follows its two-stage training process, with each subsequent stage building upon the weights learned in the previous stage. The number of views per input image is 16, and each image is encoded by a pre-trained image encoder into 576 2D image patches ~\cite{Radford2021LearningTV}, resulting in a total of 9,216 patches. Similar to mainstream open-source LMMs, LLaVA-3D is built upon the powerful generative capabilities of LLMs. It achieves multimodal understanding by aligning visual features extracted by the CLIP encoder with the text embeddings of the LMMs. The key design of LLaVA-3D lies in endowing the model with 3D spatial understanding capabilities by injecting 3D information into the 2D visual tokens.

\noindent
\textbf{Implementation Details} 
All experiments are executed on a setup comprising 4 × H20 GPUs. Our 3D evaluation benchmark primarily relies on the Scannet Datasets \cite{scannet}, where we have selected 32 scenarios and rigorously tested thousands of QA pairs within each benchmark. 
In our ablation studies, all experimental settings and test sets are kept consistent to ensure fair comparisons. We plan to disclose the specific scenarios selected later to ensure the transparency and objectivity of our experimental findings. For multi-view inputs, we apply the same masking strategy and frequency-allocation policy uniformly to every view.

\noindent
\textbf{Benchmarks} 
To comprehensively evaluate the 3D scene understanding capability of the proposed method, we conduct comprehensive experiments on two established 3D scene reasoning benchmarks: The SQA3D ~\cite{Ma2022SQA3DSQ} benchmark enriches the evaluation by incorporating dynamic 3D scenes through 19K GPT-4-generated questions. The ScanQA ~\cite{Azuma2021ScanQA3Q} dataset provides 33.4K high-quality, human-annotated question to evaluate spatial relationships and navigation planning skills.

\begin{table}[t]
\centering
\caption{Ablation study on the SQA3D subset, where we compare \textit{C}$^{2}$RoPE with multimodal image positional encoding methods CCA \cite{cca-llava} and MCA \cite{mca}. The baseline LLaVA-3D adopts the default RoPE.}
\footnotesize
\scalebox{1}{
\begin{tabular}{@{}c@{\hspace{8pt}}|cc|cc@{}} 
\toprule
 &   \multicolumn{2}{c}{\textbf{SQA3D (test) }}&\multicolumn{2}{c}{\textbf{SQA3D (val)}}\\ 
  \textit{Methods}& \multicolumn{1}{c}{\textit{EM@1}}$\uparrow$ & \multicolumn{1}{c}{\textit{EM@R}}$\uparrow$  & \multicolumn{1}{c}{\textit{EM@1}}$\uparrow$ & \multicolumn{1}{c}{\textit{EM@R}}$\uparrow$     \\ 
\midrule
 \textit{LLaVA-3D}& 55.6& 53.1& 51.5*&54.5*\\ 
  \textit{+MCA}& 55.5& 53.7&50.9&54.7\\
    \textit{+CCA}& 56.2& 53.9&50.9&54.4\\
  \midrule
 \textit{+\textit{C}$^{2}$RoPE}& 56.8& 54.3&51.6&55.7\\
\bottomrule
\end{tabular}
}
\label{cca-mca-table}
\vspace{-2em}
\end{table}

\subsection{Comparison with State-of-the-Art Methods}

As reported in Tab.~\ref{tab:mme}, compared to the baseline LLaVA-3D, our method achieves consistent improvements across all five evaluation metrics on ScanQA, with gains of +4.3 on EM@1, +8.5 on B-4, +13.4 on MET, +2.5 on RGE, and +18.1 on CIDEr. On the SQA3D test set, our method achieves improvements of +1.2 on EM@1 and +1.2 on EM@R, respectively. 

Benefiting from the proposed spatio-temporal continuous positional embedding mechanism and Chebyshev causal masking, our method significantly enhances the model's spatial reasoning and image perception capabilities. Specifically, our model achieves superior performance compared to expert models such as 3D-VLP ~\cite{3d-vlp}, which are designed for specific tasks. This is because our model can leverage the powerful multimodal understanding capabilities of LMMs to significantly enhance performance on 3D tasks. Additionally, we present the performance of 2D LLMs on ScanQA and SQA3D. Compared to these models, our method achieves overwhelming performance advantages—for instance, it surpasses Qwen2-VL-7B by +20.0 on BLEU-4 and +22.7 on METEOR. Compared to current mainstream 3D LMMs, although our baseline does not achieve the best performance on ScanQA, it attains the state-of-the-art performance after incorporating the C²RoPE proposed in this paper, further demonstrating the effectiveness of our improvements.

We observe that LLaVA-3D underperforms compared to certain 3D LMMs, such as ChatScene and Ross3D, on the SQA3D benchmark. We analyze that this is because LLaVA-3D embeds point cloud features into 2D image tokens; while this enhances the 3D representation of image tokens, the model lacks richer 3D information as input for complex spatial reasoning in intricate scenes \cite{Ma2022SQA3DSQ} (as illustrated in Figure 2). Although our method enhances the image perception of LLaVA-3D, it does not alter the way 3D information is modeled; therefore, its performance remains suboptimal compared to a few specialized 3D LMMs.

\subsection{Ablation study}

To mitigate long-term decay, CCA and MCA heuristically redirect instruction tokens to focus more on the central regions of images by reassigning different manually designed 1D positional indices to image tokens. In Fig.~\ref{cca-mca-our}, we illustrate the long-term decay under RoPE, CCA, and MCA, respectively. In Fig.~\ref{cca-mca-our}, we compare the information flow patterns of the baseline model and various improved image positional encoding methods. Our approach achieves a more balanced attention distribution across image tokens, enhancing the overall perception of visual information. Tab.~\ref{cca-mca-table} presents the quantitative results of the ablation study, where our method consistently achieves the best performance on both the validation and test subsets of SQA3D.

\subsection{Case study}

The Fig.~\ref{case} presents a case study in which both the baseline and our approach are provided with the same instruction and multi-view scene images. In the displayed example, the baseline generates a response containing hallucinated content. In contrast, our method accurately perceives the visual information and produces a correct response. 

\section{Conclusion}

This paper reveals two key limitations of RoPE, inherited from LLMs, in the context of 3D LMMs: spatial locality loss and image token neglect. Through in-depth analysis, we explain the underlying causes: 1) RoPE assigns one-dimensional temporal positional indices to image tokens, disrupting their spatial continuity along the column dimension; 2) RoPE allocates attention based on the principle that "temporally closer image tokens are more causally related," leading to long-term decay across image tokens. To address these issues, we propose C$^{2}$RoPE, which incorporates two key designs: a spatio-temporal continuous positional embedding mechanism and Chebyshev causal masking. Comprehensive experiments validate the effectiveness of C$^{2}$RoPE.

\bibliographystyle{IEEEtran}
\bibliography{IEEEexample}

\end{document}